\documentclass[letterpaper]{article} 
\usepackage{aaai2026}  
\usepackage{times}  
\usepackage{helvet}  
\usepackage{courier}  
\usepackage[hyphens]{url}  
\usepackage{graphicx} 
\usepackage{natbib}  
\usepackage{caption} 
\usepackage{multirow}
\usepackage{multicol}
\usepackage{xcolor}
\usepackage{colortbl}
\usepackage{graphicx}
\usepackage{tikz} 
\usepackage{subcaption}
\usepackage{bbding} 
\usepackage{amsmath} 
\usepackage{amsfonts}
\usepackage{times}
\usepackage{helvet}
\usepackage{courier}
\usepackage{xcolor}
\definecolor{deepgreen}{rgb}{0,0.5,0}
\usepackage{algorithm}
\usepackage{algorithmic}
\usepackage{newfloat}
\usepackage{listings}
\DeclareCaptionStyle{ruled}{labelfont=normalfont,labelsep=colon,strut=off} 
\floatstyle{ruled}
\newfloat{listing}{tb}{lst}{}
\floatname{listing}{Listing}
\newcommand{\ourmodel}{\textit{T-Rex-Omni}}

\title{\ourmodel{}: Integrating Negative Visual Prompt in Generic Object Detection}
\author{
    Jiazhou Zhou$^{1,2\dagger}$,
    Qing Jiang$^{2,3\dagger}$,
    Kanghao Chen$^{1}$, \\
    Lutao Jiang$^{1}$, 
    Yuanhuiyi Lyu$^{1}$, 
    Ying-Cong Chen$^{1}$,
    Lei Zhang$^{2\ddagger}$
}
\affiliations{

    \textsuperscript{\rm 1} Hong Kong University of Science and Technology (Guangzhou) \\
    \textsuperscript{\rm 2} International Digital Economy Academy (IDEA)\\
    \textsuperscript{\rm 3} South China University of Technology\\
    \{zhoujiazhou, jiangqing, leizhang\}@idea.edu.cn
}

\usepackage{bibentry}

\begin{document}

\maketitle
\footnotetext[1]{$\dagger$ This work was done when Jiazhou Zhou and Qing Jiang were interns at IDEA.}
\footnotetext[2]{$\ddagger$ Corresponding author.}

\begin{abstract}
Object detection methods have evolved from closed-set to open-set paradigms over the years. Current open-set object detectors, however, remain constrained by their exclusive reliance on positive indicators based on given prompts like text descriptions or visual exemplars. This positive-only paradigm experiences consistent vulnerability to visually similar but semantically different distractors. We propose \ourmodel{}, a novel framework that addresses this limitation by incorporating negative visual prompts to negate hard negative distractors. Specifically, we first introduce a unified visual prompt encoder that jointly processes positive and negative visual prompts. Next, a training-free Negating Negative Computing (NNC) module is proposed to dynamically suppress negative responses during the probability computing stage. To further boost performance through fine-tuning, our Negating Negative Hinge (NNH) loss enforces discriminative margins between positive and negative embeddings. \ourmodel{} supports flexible deployment in both positive-only and joint positive-negative inference modes, accommodating either user-specified or automatically generated negative examples. Extensive experiments demonstrate remarkable zero-shot detection performance, significantly narrowing the performance gap between visual-prompted and text-prompted methods while showing particular strength in long-tailed scenarios (51.2 AP$_r$ on LVIS-minival). This work establishes negative prompts as a crucial new dimension for advancing open-set visual recognition systems. 
\end{abstract}


\section{Introduction}
Object detection stands as a cornerstone of computer vision, tasked with precisely localizing and categorizing objects within images. The field has evolved remarkably in recent years. Initially dominated by closed-set paradigms~\cite{carion2020end,li2022dn,liu2022dabdetr,zhang2022dino,zhu2020deformable} limited to predefined categories, it has now shifted toward more flexible open-set detection systems~\cite{li2022grounded,liu2024grounding,zhou2022detecting} that can identify objects specified through user prompts, including text prompts~\cite{gu2021open,li2022grounded,liu2024grounding,zhan2024griffon,jiang2024chatrex} (``a photo of a muffin"), visual prompts~\cite{minderer2022simple,xu2023multi,zang2022open,li2024visual} (reference images of muffin), or combinations of them~\cite{jiang2024t}.


\begin{figure}[t]
\centering
\includegraphics[width=0.44\textwidth]{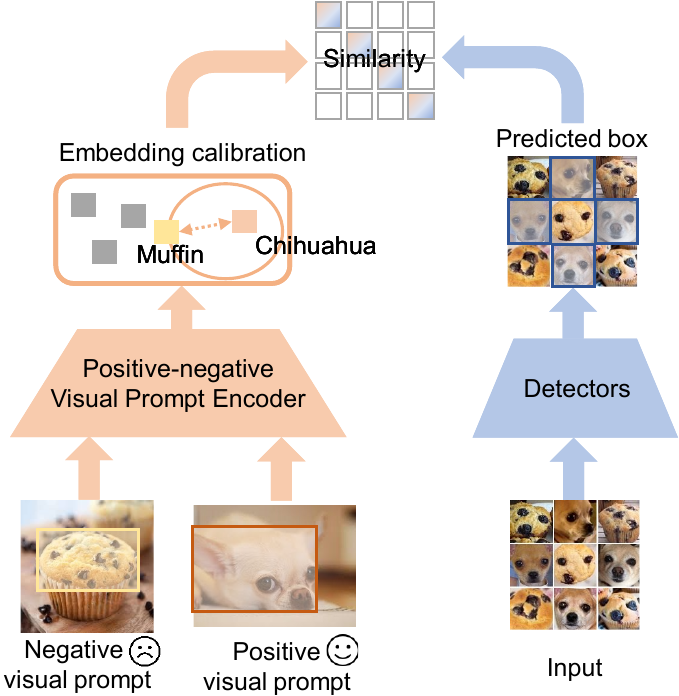}
\caption{\ourmodel{} employs dual visual prompts to enhance detection precision: positive prompts (e.g., ``Chihuahua") guide target object localization while negative prompts (e.g., ``muffin") actively suppress visually similar distractors. The joint positive-negative framework enables more discriminative and user-specified object detection.}
\label{fig:Tesear}
\end{figure}

Modern open-set detection systems, however, face a fundamental limitation: their exclusive reliance on positive indicators from given prompts leaves them vulnerable to hard negatives—visually similar yet semantically distinct instances. As illustrated in Fig.~\ref{fig:Tesear}, even state-of-the-art detectors may confidently classify a Chihuahua as a muffin when relying solely on positive visual prompts (e.g., a bounding box example of a Chihuahua). This issue is exacerbated by long-tailed data distributions, where detectors underperform for rare categories, leading to poor real-world applicability. To mitigate this, we propose leveraging negative visual prompts (e.g., a bounding box example of a muffin) to explicitly guide detectors away from hard negative distractors while preserving sensitivity to positive instances. This motivates our core research question:
\textit{Can visual negative prompts enable models to actively negate hard negatives without compromising their ability to detect true positives?}

Our solution, \ourmodel{}, systematically integrates negative prompts into modern detection frameworks. First, we introduce \textbf{a unified positive-negative prompt encoder} that jointly processes positive and negative visual prompts from single or multiple images into corresponding prompt embeddings. To mitigate data scarcity, our prompt encoder synthesizes visual prompts by randomly jittering and resizing ground-truth boxes (mild for positives, strong for negatives). This augmentation enhances reference robustness to spatial variations and supports cross-image object detection. Next, we propose \textbf{a Negating Negative Computing (NNC) module}, which adaptively suppresses negative responses during probability computation. This training-free design permits immediate deployment with significant performance gains (Tab.~\ref{tab:main_ab}). For further improvement via fine-tuning, \textbf{a Negating Negative Hinge (NNH) loss} enforces discriminative margins between positive and negative prompts in the embedding space, actively pushing apart visually similar but semantically distinct category embeddings (Tab.~\ref{tab:main_ab}).

In this way, \ourmodel{} offers three flexible visual prompt settings during inference: \textbf{(1) User-curated mode:} Users explicitly specify both positive and negative exemplars for precision-critical applications. \textbf{(2) Auto-suggested mode:} The system automatically proposes relevant negative exemplars based on user-provided positives, enabling efficient deployment with minimal user input. \textbf{(3) Positive-only mode:} Traditional single-prompt operation for rapid deployment. \ourmodel{} tri-mode inference enables practitioners to dynamically adapt to varying precision and efficiency requirements across different application scenarios.

\ourmodel{} exhibits strong object detection performance across four challenging benchmarks (COCO, LVIS, ODinW, Roboflow100) in zero-shot settings. Our key findings reveal that negative visual prompts can mitigate the previous modality gap between text-prompt and visual-prompt methods. \ourmodel{} (Swim-L) even surpasses traditional text-prompt methods by \textbf{+2.0 AP} (LVIS-val) in Tab.~\ref{tab:main_result}. Besides, it delivers exceptional performance for rare categories on LVIS-minival (\textbf{51.2 AP$_r$}), LVIS-Val  (\textbf{49.8 AP$_r$}), ODinW (\textbf{29.6 AP$_{avg}$}), and Roboflow (\textbf{20.3 AP$_{avg}$}), significantly outperforming existing approaches in long-tailed scenarios (Tab.~\ref{tab:main_result}). To summarize, our contributions are threefold:
\begin{itemize}
    \item A simple yet effective framework that integrates negative examples as prompts for object detection, which can negate hard negative distractors while maintaining sensitivity to positive examples.
    \item An approach for embedding space modification through negative examples, supported by the plug-and-play NNC module to suppress hard negative probabilities in a training-free manner and the NNH loss for enforced positive and negative embedding separation for fine-tuning adaptation.
    \item Extensive empirical validation across multiple benchmarks (COCO, LVIS, ODinW, Roboflow 100) demonstrating consistent zero-shot improvements, particularly in challenging long-tailed scenarios.
\end{itemize}

\section{Related Work}
\label{sec:related_work}

\noindent \textbf{Object Detection} Object detection (OD) has transitioned from closed-set to open-set approaches to fit the dynamic and unpredictable nature of real-world environments. Early closed-set detection systems~\cite{carion2020end,li2022dn,liu2022dabdetr,zhang2022dino,zhu2020deformable} are limited to recognizing objects from predefined categories. By contrast, open-set models adapt to identify objects beyond initial training categories. A prevalent method for open-set object detection leverages \textbf{text prompts}~\cite{gu2021open,liu2024grounding,yao2022detclip}, which typically leverage knowledge from language models like CLIP~\cite{radford2021learning} or BERT~\cite{devlin2019bert} to align textual descriptions with visual representations. Recent multimodal large language models (MLLMs)~\cite{hurst2024gpt,bai2025qwen2,wu2024deepseek} have enhanced models with text referring expressions to capture object attributes, relationships, spatial configurations, and their interactions~\cite{jiang2024chatrex,chen2023shikra,ma2024groma}. However, texts often fail to accurately describe objects that are difficult to articulate. The \textbf{visual prompt-based} model provides a more intuitive representation through visual examples, such as image-level prompts (raw images) and object-level prompts (boxes, points). In this paper, we focus on the visual prompt-based object detection, which can benefit from open-vocabulary long-tailed objects.

\noindent \textbf{Negative Sampling} Negative sampling selects informative negatives or generates synthetic ones to improve representation learning while maintaining computational efficiency~\cite{duan2024negating}. This technique has demonstrated broad applicability across domains including recommendation systems~\cite{yang2020mixed,shi2023theories}, natural language processing~\cite{yang2024trisampler,zhan2021optimizing}, graph learning~\cite{yang2020understanding,duan2022learning}, and computer vision~\cite{yang2024does,wang2021exploring}. Negative sampling plays a pivotal role in object detection. Focal Loss~\cite{lin2017focal} dynamically up-weights hard negatives during training. NP-RepMet~\cite{yang2020restoring} jointly optimizes negative and positive prototypes for few-shot detection. UNP~\cite{yan2024understanding} isolates confusing negatives while ensuring the contribution of hard negatives via gradient modulation. GenNeg~\cite{zhao2024generating} leverages large-language models and text-to-image diffusion models to synthesize negative object descriptions and images. While these methods demonstrate the value of negative information for object detection, they are fundamentally constrained by their limited generalization capability for unseen categories due to their inherent dependence on training-time optimization. \ourmodel{} introduces a fundamental shift by: \textbf{(1)} enabling dynamic specification of negative samples using bounding box visual prompts; \textbf{(2)} supporting both training-free immediate deployment and fine-tunable versions for enhanced performance; and \textbf{(3)} excelling in long-tailed scenarios where previous methods struggle.

\begin{figure*}[t]
\centering
\includegraphics[width=1.0\textwidth]{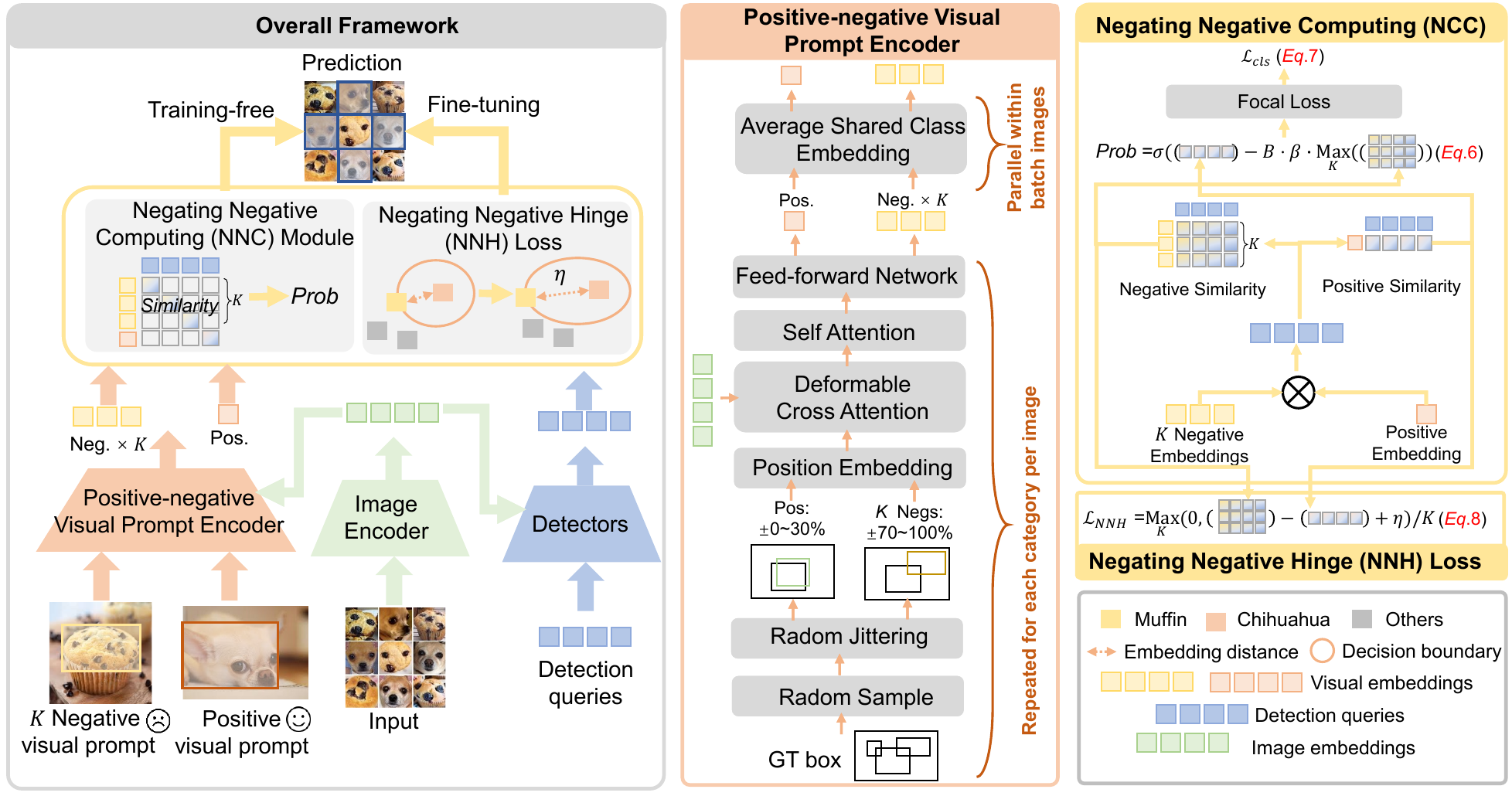} 
\caption{Overview of the \ourmodel{} model.}
\label{fig:main_framework}
\end{figure*}

\section{Model}
\subsection{Preliminary and Overview}

\noindent \textbf{Preliminary for \textit{T-Rex2}.}
\textit{T-Rex2}~\cite{jiang2024t} adopts the DETR~\cite{carion2020end} framework, implementing an end-to-end Transformer-based architecture for open-set object detection.  The model supports both joint and independent use of visual and text prompts through four key components: (1) a text prompt encoder based on CLIP's text encoder, (2) a visual prompt encoder employing deformable cross-attention to process box/point interactions, (3) an image encoder for feature extraction, and (4) a DETR-style decoder for detection. \textit{T-Rex2} enables generic and interactive visual prompt workflows for open-set object detection.

\noindent \textbf{Overview for \ourmodel{}.}
Built upon the \textit{T-Rex2} architecture~\cite{jiang2024t}, \ourmodel{} streamlines the framework by removing the text-prompt branch while introducing three novel components: (1) a unified prompt encoder for joint positive-negative prompt processing, (2) a training-free Negating Negative Computing (NNC) module for dynamic probability calibration, and (3) a Negating Negative Hinge (NNH) loss for discriminative embedding separation. As shown in Fig.~\ref{fig:main_framework}, we preserve \textit{T-Rex2}'s image encoder and DETR-style decoder, enhancing them with our proposed modules for negative prompt integration. The following subsections detail these contributions, with ablation studies in the experiment section validating their impacts.

\subsection{Positive-negative Visual Prompt Encoder}
\ourmodel{} aims to learn both positive and negative visual prompt embeddings and integrate them into object detection. Building upon \textit{T-Rex2}'s framework, we present several modifications to its original visual prompt encoder.

\noindent \textbf{Visual Prompt Generation.} During training, we generate both positive and negative visual prompts by performing random sampling and geometric jittering of the ground truth annotations. Specifically, for each category $c$ present in an image, we first randomly sample one ground truth bounding box, denoted as $G_{c} = (x_{c}, y_{c}, w_{c}, h_{c})$. Then, positive visual prompts $p_{c}$ are synthesized by applying mild transformations, namely, random scaling or shifting within $[0,0.3]$ scale range for the sampled ground truth bounding box $G_{c}$. Similarly, multiple negative visual prompts $n_{c}^{i}, i \in {1,2,...,K}$ are generated with stronger transformations within $[0.7, 1.0]$ scale range by repeating this process $K$ times. This augmentation enriches prompt diversity while preserving semantic validity. Besides, it enhances robustness to spatial and scale variations in test-time visual prompts. 

During inference, we maintain the user-provided positive visual prompt without modification. The negative visual prompts are obtained via consistent training augmentation (auto-suggested mode) or specified by users (user-curated mode).

\noindent \textbf{Visual Prompt Encoder.} 
The visual prompt encoder transforms both positive and negative visual prompts from coordinate space to embedding space. Given positive visual prompts $p_{c}$ and $K$ negative visual prompts $n_{c}^{i}, i \in {1,2,...,K}$, we initialize two learnable prompt queries $Q_P \in \mathbb{R}^{1 \times D}$ and $Q_N \in \mathbb{R}^{K \times D}$, where $D$ is the dimension. Given the multi-scale image features $F=\{f_{j} | j\in {1,2,...,L}\}$ extracted from the image encoder with $L$ feature map layers, we compute the enhanced queries via multi-scale deformable cross-attention~\cite{zhu2020deformable}. The positive prompt queries $Q_{P}^{'}$ and negative prompt queries $Q_{N}^{'}$ are computed as:


\begin{equation}
    Q_{P}^{'} = MSDeformAttn(Q_{P},p_{c},F),
\end{equation}
\begin{equation}
    Q_{N}^{'} = MSDeformAttn(Q_{N},n_{c},F),
\end{equation}
The deformable attention allows each prompt query to dynamically attend to the most relevant image features within its respective visual prompt region. These attended features are then refined through a self-attention layer followed by a feed-forward network (FFN). The positive $V_{P}$ and negative $V_{P}$ prompt embeddings are obtained as follows:
\begin{equation}
    V_{P} = FFN(SelfAttn(Q_{P}^{'})), 
\end{equation}
\begin{equation}
    V_{N} = FFN(SelfAttn(Q_{N}^{'})),
\end{equation}
The above operations are repeated across all categories for each image. Given a training batch of $B$ images with a maximum of $M$ categories per batch, we denote the positive prompts as $V^{'}_{P} \in \mathbb{R}^{M\times B\times D}$ and $K$ negative visual prompts
as $V^{'}_{N} \in \mathbb{R}^{M\times B\times  K\times D}$.

To facilitate \ourmodel{} for cross-image object detection, we enforce that each training batch contains at least one shared category (see Sec. 4.1). This enables us to compute positive averaged prompt embeddings of shared categories across the batch, namely $V^{''}_{P}\in \mathbb{R}^{M\times D}$, thus effectively propagating prompt embeddings between images of the same category. For negative prompt embeddings, we select the top-K most similar embeddings to the averaged positive embedding $V^{''}_{P}$ from the full set of $B \times K$ candidate negatives, yielding the final negative embeddings $V^{''}_{N}\in \mathbb{R}^{M\times K\times D}$. The above batch-wise operation is disabled during inference. Overall, the $V^{''}_{P}$ captures the target object's visual characteristics while $V^{''}_{N}$ represents visually similar but incorrect or suboptimal detections that should be distinguished from the target.

\noindent \textbf{Flexible Inference.} During deployment, \ourmodel{} supports three visual prompt settings to accommodate diverse application requirements: \textbf{(1) User-curated mode:} Users explicitly specify both positive and negative exemplars through bounding box annotations or precise image crops. This high-precision mode is particularly valuable for critical applications. \textbf{(2) Auto-suggested mode:} The system automatically proposes relevant negative exemplars by applying geometric transformations to user-provided positive exemplars. This balanced mode reduces annotation effort and serves as our default setting for benchmark evaluations. \textbf{(3) Positive-only mode:} Maintains compatibility with traditional positive-only prompt workflows for rapid deployment scenarios. \ourmodel{} tri-mode inference enables practitioners to dynamically adapt to varying precision and efficiency requirements across different application scenarios.

\subsection{Negating Negative Computing Module.}
Built upon the extracted positive $V^{''}_{P}$ and $K$ negative prompt embeddings $V^{''}_{N,i}, i=1,2,...,K$, our NNC module calibrates prediction confidence by suppressing scores for hard negative cases. As shown in Fig.~\ref{fig:main_framework}, given $N_{q}$ detection queries $Q\in \mathbb{R}^{N_{q} \times D_{q}}$ from DETR decoder, we compute positive $S_{P}$ and negative $S_{N,i}$ similarity scores between detection queries with both positive and multiple negative prompt embeddings via matrix multiplication:
\begin{equation}
S_{P}, S_{N,i} = Q \times (V^{''}_{P})^{T}, Q \times (V^{''}_{N,i})^{T}
\end{equation}
We then subtract the positive similarity scores from the weighted negative ones to suppress the likelihood of predicting classes that are visually similar but semantically different. The following sigmoid function $\sigma$ then transforms the similarity score into probability $Prob$ as follows:
\begin{equation}
    Prob = \sigma(S_{P} - B \cdot \beta \cdot \mathop{\max}_{i=1,2,...,K}(S_{N,i})),
\end{equation}

where $0<\beta<1$ is the introduced parameter to control how much the negative examples influence the final score; the max operation $\text{Max}(\cdot)$ selects the strongest negative similarity across $K$ negative similarities and $B \sim \text{Bernoulli}(0.5)$ is a stochastic indicator for mode switching. During training, $B$ stochastically switches between joint positive-negative ($B=1$) and positive-only ($B=0$) modes to ensure inference compatibility. 

In training-free or inference applications, NNC operates as a plug-and-play module, demonstrating consistent performance without fine-tuning (Tab.~\ref{tab:main_ab}). For fine-tuning training applications, we integrate the predicted probabilities from the NNC module into Focal loss~\cite{lin2017focal} to calculate the classification loss and backpropagation:
\begin{equation}
    \mathcal{L}_{cls} = -\alpha_t (1 - Prob_t)^\gamma \log(Prob_t),
\end{equation}
where $Prob_t$ is defined as $Prob_t=Prob$ if the class prediction is true, otherwise $Prob_t=1-Prob$; $\alpha_t \in [0,1]$ is the class-balancing weight and $\gamma \geq 0$ controls the focus on hard examples.

\subsection{Negating Negative Hinge Loss.}
To improve discrimination between visually similar but semantically distinct categories, we propose the Negating Negative Hinge (NNH) loss, which explicitly enforces a margin-based separation between positive and negative prompt embeddings. The loss is defined as:
\begin{equation}
    \mathcal{L}_{Hinge} = \sum_{i=1,2,...,K}\text{Max}(0,S_{N,i}-S_{P}+\eta) / K
\end{equation}
where $S_{P}$ and $S_{N,i}$ are the similarity scores for the positive and the \textit{i}-th negative prompt embedding; $\eta > 0$ is a preset margin that controls the minimum separation between positive and negative similarities; and $K$ is the number of negative prompt embeddings.

The NNH loss ensures that the similarity $S_{P}$ for the positive embeddings exceeds the similarity $S_{N,i}$ for any negative embeddings by at least the margin $\eta$. This constraint encourages the model to learn more discriminative embeddings by penalizing cases where negative similarities overlap with positive similarities. The hinge loss formulation $\text{Max}(0, \cdot)$ provides a robust optimization objective, as it only penalizes violations of the margin condition, making the training process more stable and focusing on hard negative cases.

\subsection{Training Strategy and Objective.}
\noindent \textbf{Training Strategy.} Unlike \textit{T-Rex2}~\cite{jiang2024t}, which employs a ``current image prompt, current image detect'' training paradigm, we introduce a ``current image prompt, cross-image detect'' training strategy. By ensuring each training batch contains at least one shared category across images (see Sec. 4.1), our approach encourages more robust visual prompt learning through inter-image consistency, thus enhancing cross-image object detection capability and generalization of prompt embeddings.

\noindent \textbf{Training Objectives.} Our complete loss function combines box regression losses (L1 and GIoU~\cite{rezatofighi2019generalized}), classification loss $\mathcal{L}_{cls}$ (\textit{Eq}.~9), our NNH loss $\mathcal{L}_{Hinge}$ (\textit{Eq}.~10), auxiliary losses (intermediate supervision after each decoder layer and encoder outputs), and denoising training loss proposed in DINO~\cite{zhang2022dino} to accelerate convergence. The box regression and classification loss are initially employed for bipartite matching~\cite{carion2020end} between predictions and ground truths. The final objective function is:
\begin{equation}
    \mathcal{L}_{total} = \mathcal{L}_{cls}+\mathcal{L}_{Hinge}+\mathcal{L}_{L1}+\mathcal{L}_{GIoU}+\mathcal{L}_{DN}
\end{equation}

\begin{table*}[h]
\setlength{\tabcolsep}{1mm}
  \resizebox{\linewidth}{!}{
    \begin{tabular}{cc|c|cccccccc|cc|c}
      \cline{1-14}
      \multirow{3}{*}{Methods} & \multirow{3}{*}{\begin{tabular}[c]{@{}c@{}}Prompt \\ Type\end{tabular}} & \begin{tabular}[c]{@{}c@{}}COCO-Val\\ Zero-Shot\end{tabular} & \multicolumn{8}{c|}{\begin{tabular}[c]{@{}c@{}}LVIS \\ Zero-Shot\end{tabular}} & \multicolumn{2}{c|}{\begin{tabular}[c]{@{}c@{}}ODinW35\\ Zero-Shot\end{tabular}} & \begin{tabular}[c]{@{}c@{}}Roboflow100\\ Zero-Shot\end{tabular} \\ \cline{3-14} 
      &   & val-80 & \multicolumn{4}{c|}{minival-804} & \multicolumn{4}{c|}{val-1203} & \multicolumn{2}{c|}{val-35} & val-100 \\\cline{3-14} 
      &  & AP & AP & AP$_f$ & AP$_c$ & \multicolumn{1}{c|}{$AP_{r}$} & AP & AP$_f$ & AP$_c$ & AP$_r$ & AP$_{avg}$ & AP$_{med}$ & AP$_{avg}$ \\ \cline{1-14}
      
      \multicolumn{14}{l}{\textbf{Swin-T Backbone}} \\ \cline{1-14}
      GLIP-T~\cite{li2022grounded} & Text & 46.7 & 26.0 & 31.0 & 21.4 & \multicolumn{1}{c|}{20.8} & 17.2 & 25.5 & 12.5 & 10.1 & 19.6 & 5.1 & - \\
      Grounding DINO~\cite{liu2023grounding} & Text & 48.4 & 27.4 & 32.7 & 23.3 & \multicolumn{1}{c|}{18.1} & - & - & - & - & 22.3 & 11.9 & - \\
      DetCLIPv2~\cite{yao2023detclipv2} & Text & - & 40.4 & 40.0 & 41.7 & \multicolumn{1}{c|}{36.0} & - & - & - & - & - & - & - \\
      MM-GDINO~\cite{zhao2024open} & Text & - & 41.4 & 46.2 & 37.4 & \multicolumn{1}{c|}{34.2} & 31.9 & 40.5 & 27.6 & 23.6 & 23.1 & - & - \\
    T-Rex2~\cite{jiang2024t} & Text & 45.8 & 42.8 & 46.5 & 39.7 & \multicolumn{1}{c|}{37.4} & 34.8 & 41.2 & 31.5 & 29.0 & 18.0 & 4.7 & 8.2 \\
    LLMDet~\cite{fu2025llmdet} & Text & - & 44.7 & 50.7 & 39.5 & \multicolumn{1}{c|}{37.3} & 34.9 & 44.3 & 30.1 & 26.0 & 23.8 & - & - \\\hline
    DINOv~\cite{li2023visual} & Visual-G & - & - & - & - & \multicolumn{1}{c|}{-} & - & - & - & - & 14.9 & 5.4 & - \\ 
      T-Rex2~\cite{jiang2024t} & Visual-G & 38.8 & 37.4 & 41.8 & 33.9 & \multicolumn{1}{c|}{29.9} & 34.9 & 41.1 & 30.3 & 32.4 & 23.6 & 17.5 & 17.4 \\
    VisTex-DINO~\cite{wu2025visual} & Text+Visual-G & - & 42.8 & - & - & \multicolumn{1}{c|}{37.2} & - & - & - & - & - & - & - \\
       \ourmodel{} & Visual-G & 43.6 & 43.0 & 47.7 & 38.9 & \multicolumn{1}{c|}{37.0} & 37.7 & 41.9 & 33.6 & 38.6 & 25.2 & 20.1 & 18.9 \\ \cline{1-14}
      
      \multicolumn{14}{l}{\textbf{Swin-L Backbone}} \\ \cline{1-14}
      GLIP-L~\cite{li2022grounded} & Text & 49.8 & 37.3 & 41.5 & 34.3 & \multicolumn{1}{c|}{28.2} & 26.9 & 35.4 & 23.3 & 17.1 & 23.4 & 11.0 & 8.6 \\
      Grounding DINO~\cite{liu2023grounding} & Text & 52.5 & 33.9 & 38.8 & 30.7 & \multicolumn{1}{c|}{22.2} & - & - & - & - & 26.1 & 18.4 & - \\
      DetCLIPv2~\cite{yao2023detclipv2} & Text & - & 44.7 & 43.7 & 46.3 & \multicolumn{1}{c|}{43.1} & - & - & - & - & - & - & - \\
      MM-GDINO~\cite{zhao2024open} & Text & - & 36.8 & 42.8 & 31.8 & \multicolumn{1}{c|}{28.1} & 29.1 & 37.2 & 25.6 & 19.7 & - & - & - \\
      LLMDet~\cite{fu2025llmdet} & Text & - & 51.1 & 56.6 & 46.1 & \multicolumn{1}{c|}{45.1} & 42.0 & 50.2 & 38.8 & 31.6 & - & - & - \\
    T-Rex2~\cite{jiang2024t} & Text & 52.2 & 54.9 & 56.1 & 54.8 & \multicolumn{1}{c|}{49.2} & 45.8 & 50.2 & 43.2 & 42.7 & 22.0 & 7.3 & 10.5 \\ \hline
    DINOv~\cite{li2023visual} & Visual-G & - & - & - & - & \multicolumn{1}{c|}{-} & - & - & - & - & 15.7 & 4.8 & - \\
      T-Rex2~\cite{jiang2024t} & Visual-G & 46.5 & 47.6 & 49.5 & 46.0 & \multicolumn{1}{c|}{45.4} & 45.3 & 49.5 & 42.0 & 43.8 & 27.8 & 20.5 & 18.5 \\
    VisTex-GLIP~\cite{wu2025visual} & Text+Visual-G & - & 50.7 & - & - & \multicolumn{1}{c|}{42.9} & - & - & - & - & - & - & - \\
       \ourmodel{} & Visual-G & \textbf{50.7} & \textbf{54.0} & \textbf{56.0} & \textbf{52.4} & \multicolumn{1}{c|}{\textbf{51.2}} & \textbf{47.8} & \textbf{50.0} & \textbf{45.1} & \textbf{45.1} & \textbf{29.6} & \textbf{23.1} & \textbf{20.3} \\ \cline{1-14}
    \end{tabular}}
\caption{Zero-shot object detection results. Best visual-prompted results are marked in \textbf{bold}.}
\label{tab:main_result}
\end{table*}

\section{Experiments}
\subsection{Data Engine}
\label{sub:Data_Engines}
Our visual prompt object detection framework employs a specialized batch construction strategy to enable cross-image detection. Each training batch contains images sharing at least one object category, allowing object instances from one image to serve as visual prompts for detecting corresponding instances in other batch images. The batch construction involves two key steps: \textbf{(1)} building a hash table that maps object categories to images containing more than three instances to ensure instance diversity, and \textbf{(2)} for each image, selecting its second-most frequent category and retrieving matching images from the hash table. We apply this strategy to generate training batches for fine-tuning on the Objects365 dataset~\cite{shao2019objects365}.

\subsection{Model Details}
\ourmodel{} adopts the pre-trained weights from T-Rex2~\cite{jiang2024t}, featuring a Swin Transformer~\cite{liu2021swin} backbone and six-layer Transformer encoder. Designed specifically for visual-prompt object detection, our architecture omits text encoders and employs: \textbf{(1)} a visual prompt encoder with three deformable cross-attention layers with a hidden dimension set to 1024 and \textbf{(2)} a prompt sampling strategy that selects one ground-truth instance box as a positive prompt and three randomly jittered boxes as negative prompts per category. 
The hyperparameters $\beta$ for the NNC module, $\eta$ for the NNH loss, $\alpha_t$ and  $\gamma$ for the focal loss are 0.3, 0.3, 0.25, and 2, respectively. 
We optimize using AdamW~\cite{loshchilov2017decoupled} with differential learning rates ($10^{-5}$ backbone, $10^{-4}$ others). 
The batch size is six, where each batch shares at least one common object category to facilitate cross-image detection.

\subsection{Settings and Metrics}
We conduct zero-shot evaluation where the training images used by \ourmodel{} has no overlap with the evaluation dataset. We report Average Precision (AP) metrics across four benchmarks: COCO~\cite{lin2014microsoft}, LVIS~\cite{gupta2019lvis}, ODinW35~\cite{li2022elevater}, and Roboflow100~\cite{ciaglia2022roboflow}. We employ the following visual prompt settings for evaluation:

\noindent \textbf{Visual-G}: In this setting, we adhere to \textit{T-Rex2}~\cite{jiang2024t} Visual-G evaluation protocol for open-set object detection. For each benchmark category, we extract both positive and negative visual prompt embeddings from the training set images, with negatives created by randomly jittering ground truth boxes. Taking COCO as a representative example, this visual prompt embedding generation process follows three steps: \textbf{(1) Sampling:} For each category, we randomly select $N=16$ images containing at least one instance of that category from the training dataset. \textbf{(2) Embedding Extraction:} We extract positive and three negative visual embeddings using each image's ground truth box and corresponding jittered boxes as input, respectively. \textbf{(3) Aggregation:} We compute category-level average embeddings, thus yielding 80 positive embeddings and $80 \times K$ negative embeddings for COCO, where we set $K=3$ in this evaluation based on our ablation study (Fig.~\ref{fig:ab_hyperparameters}d). This process is performed once and remains fixed during evaluation, ensuring consistent prompt representation across the benchmark.

\subsection{Main Results}
\noindent \textbf{Zero-Shot Generic Object Detection.}
We conduct a comprehensive evaluation of \ourmodel{}'s zero-shot capabilities (denoted as ``Visual-G" in Tab.~\ref{tab:main_result}) across four challenging benchmarks, where ``zero-shot" denotes evaluation on images excluded from training. Our analysis reveals four key findings: \textbf{(1) State-of-the-Art Visual Prompting:} With Swin-T, \ourmodel{} outperforms the previous best visual-prompt approach (T-Rex2) by \textbf{+4.8 AP} (43.6 vs 38.8) on COCO-val and \textbf{+5.6 AP} (43.0 vs 37.4) on LVIS-minival, setting new visual-prompt benchmarks. \textbf{(2) Long-Tailed Superiority:} The most striking improvement appears in long-tailed scenarios. \ourmodel{} achieves \textbf{+7.1 AP$_r$} for LVIS-minival rare categories (37.0 vs 29.9), demonstrating 23.8\% relative improvement in long-tailed scenarios. \textbf{(3) Text-Visual Gap Reduction}: \ourmodel{} narrows the text-visual performance gap to \textbf{2.2 AP} on COCO (43.6 vs T-Rex2-text's 45.8) while surpassing text-prompt T-Rex2 on LVIS-val by \textbf{+2.8 AP} (37.7 vs 34.9). \textbf{(4) Backbone Scalability}: With the Swin-L backbone, improvements remain consistent: \textbf{+4.2 AP} on COCO (50.7 vs 46.5) and \textbf{+5.8 AP$_r$} for LVIS rare categories (51.2 vs 45.4), with ODinW-35 gains of \textbf{+1.8 AP$_{avg}$}.

\subsection{Ablation Experiments}
\label{sec:Ablation}

\begin{table}[]
\setlength{\tabcolsep}{0.7mm}
\resizebox{0.47\textwidth}{!}{
\begin{tabular}{ccc|c|cccc}
\hline
\multicolumn{3}{c|}{Settings} & \begin{tabular}[c]{@{}c@{}}COCO-val \\ Zero-Shot\end{tabular} & \multicolumn{4}{c}{\begin{tabular}[c]{@{}c@{}}LVIS-minival \\ Zero-Shot\end{tabular}} \\ \hline
 NNC & NNH & Fine-tune & AP & AP & AP$_r$ & AP$_c$ & AP$_f$ \\ \hline
\XSolidBrush & \XSolidBrush & \XSolidBrush & 38.8 & 37.4 & 41.8 & 33.9 & 29.9 \\
\checkmark & \XSolidBrush & \XSolidBrush & 41.8 {\scriptsize +3.0} & 40.6 {\scriptsize +3.2} & 45.2 {\scriptsize +3.4} & 36.9 {\scriptsize +3.0} & 33.3 {\scriptsize +3.4} \\
\checkmark & \XSolidBrush & \checkmark & 42.9 {\scriptsize +4.1} & 41.4 {\scriptsize +4.0} & 46.2 {\scriptsize +4.4} & 38.1 {\scriptsize +4.2} & 35.1 {\scriptsize +5.2} \\
\checkmark & \checkmark & \checkmark &\textbf{43.6} {\scriptsize +4.8} & \textbf{43.0} {\scriptsize +5.6} & \textbf{47.7} {\scriptsize +5.9} & \textbf{38.9} {\scriptsize +5.0} & \textbf{37.0} {\scriptsize +7.1} \\
\hline
\end{tabular}}
\caption{Ablation study on the NNC module and NNH loss in fine-tuning and training-free modes.}
\vspace{-5pt}
\label{tab:main_ab}
\end{table}

\noindent \textbf{Ablation Study on the NNC Module and NNH Loss.}
As shown in Tab.~\ref{tab:main_ab}, our NNC module significantly improves performance in a training-free setting (row 2), increasing COCO-val AP by \textbf{+3.0} and LVIS-minival AP by \textbf{+3.2}. This validates NNC’s probability calibration (Eq. 6) in suppressing hard negative predictions. Further fine-tuning (row 3 using Eq. 9) enhances performance (COCO-val: \textbf{+1.1}, LVIS-minival: \textbf{+0.8}) while the full model (row 4), which incorporates the NNH loss (Eq. 8), achieves the best results (COCO-val: \textbf{43.6} AP, LVIS-minival: \textbf{43.0} AP), demonstrating the NNH loss's ability to enforce discriminative margins in embedding space (Eq. 8). These results validate NNC’s stochastic negative suppression and fine-tuning jointly reduce false positives, while NNH’s margin separation enhances inter-class positive-negative discrimination.

\begin{table}[]
\setlength{\tabcolsep}{1.8mm}
\resizebox{0.47\textwidth}{!}{
\begin{tabular}{cc|c|cccc}
\hline
\multicolumn{2}{c|}{Settings} & \begin{tabular}[c]{@{}c@{}}COCO-val\\ Zero-Shot\end{tabular} & \multicolumn{4}{c}{\begin{tabular}[c]{@{}c@{}}LVIS-minival\\ Zero-Shot\end{tabular}} \\ \hline
\multicolumn{1}{c|}{Training} & \multicolumn{1}{c|}{Evaluation} & AP & AP & AP$_r$ & AP$_c$ & AP$_f$ \\ \hline
\multicolumn{1}{c|}{\multirow{2}{*}{$B=0$}} & \multicolumn{1}{c|}{$B=0$}  & 38.8 & 37.4 & 41.8 & 33.9 & 29.9 \\
\multicolumn{1}{c|}{} & \multicolumn{1}{c|}{$B=1$}  & 41.8  & 40.6 & 45.2 & 36.9 & 33.3 \\ \hline
\multicolumn{1}{c|}{\multirow{2}{*}{$B=1$}} & \multicolumn{1}{c|}{$B=0$}  & 40.8 & 39.8 & 44.0 & 35.7 & 31.1 \\
\multicolumn{1}{c|}{} & \multicolumn{1}{c|}{$B=1$}  & 42.4  & 40.9 & 45.5 & 37.5 & 34.2 \\ \hline
\multicolumn{1}{c|}{\multirow{2}{*}{\begin{tabular}[c]{@{}c@{}} $B(0.5)$ \end{tabular}}} & \multicolumn{1}{c|}{$B=0$} & 42.0 & 41.0 & 45.9 & 37.2 & 34.8 \\
\multicolumn{1}{c|}{} & \multicolumn{1}{c|}{$B=1$}  & \textbf{43.6}  & \textbf{43.0} & \textbf{47.7} & \textbf{38.9} & \textbf{37.0} \\ \hline
\end{tabular}
}
\caption{Ablation study on \ourmodel{}'s compatibility for positive-only ($B=0$) and positive-negative ($B=1$) visual prompt settings. $B \sim \text{Bernoulli}(0.5)$ is an indicator for stochastic mode switching.} 
\label{tab:positive-only and positive-negative}
\vspace{-5pt}
\end{table}

\noindent \textbf{Ablation Study on Prompt Setting Compatibility.}
Our systematic evaluation (Tab.~\ref{tab:positive-only and positive-negative}) reveals three key findings regarding the mode switching training mechanism. First, models trained with fixed negative prompt integration ($B=1$) show better performance in positive-negative evaluation (\textbf{42.4} AP on COCO-val) than in positive-only mode (\textbf{40.8} AP on COCO-val $\Delta=+1.6$), confirming the value of negative prompt utilization. Second, stochastic training yields more robust performance across both evaluation settings, achieving better results in the positive-negative configuration (\textbf{43.6} AP on COCO-val, \textbf{+2.8} over fixed training). Most notably, the approach demonstrates exceptional performance on rare categories of LVIS-minival (\textbf{47.7} AP$_r$), with a \textbf{6.9} point improvement over fixed training, highlighting its effectiveness for long-tailed recognition. These consistent gains across COCO-val and LVIS-minival demonstrate that our mode switching training mechanism successfully bridges the gap between different prompt configurations while maintaining superior discriminative capabilities.

\begin{figure*}[h]
\centering
\includegraphics[width=1.0\textwidth]{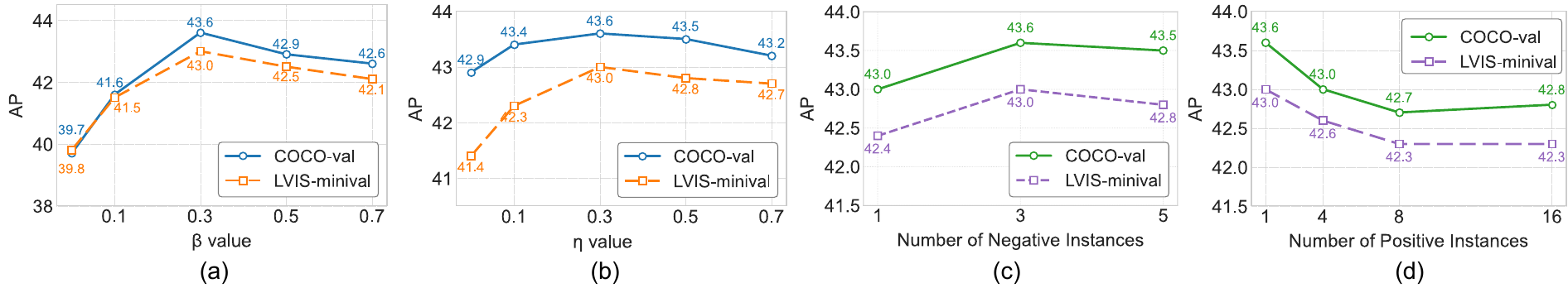}
\caption{Ablation study on hyperparameters. (a) $\beta$ in the NNC module; (b) $\eta$ in the NNH loss; (c) positive prompt quantity; (d) negative prompt quantity.}
\label{fig:ab_hyperparameters}
\end{figure*}

\begin{figure*}[h]
\centering
\includegraphics[width=1.0\textwidth]{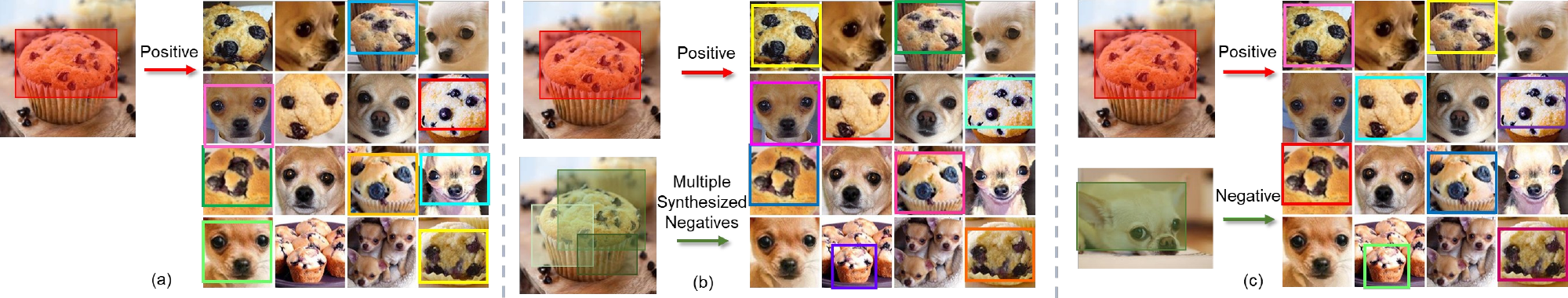}
\caption{Visualization of \ourmodel{}'s three inference modes. (a) Positive-only; (b) Auto-suggested; (c) User-curated.}
\label{fig:ab_visulization}
\end{figure*}


\noindent \textbf{Hyperameter $\beta$ Selection in the NNC module.} We ablate the negative suppression coefficient $\beta$ in the NNC module (Fig.~\ref{fig:ab_hyperparameters}a). Results show a unimodal performance curve, peaking at $\beta=0.3$ (42.8 AP on COCO-val, 43.0 AP on LVIS-minival), balancing positive reinforcement and negative suppression. Performance degrades at extremes: no suppression ($\beta=0.0$) yields 39.7 AP, while over-suppression ($\beta\geq0.5$) causes decline. This confirms that moderate suppression ($\beta\approx0.3$) is critical—both neglecting and over-penalizing negatives harm discriminative power.

\noindent \textbf{Hyperparameter $\eta$ Selection in the NNH loss.} As shown in Fig.~\ref{fig:ab_hyperparameters}b, the optimal detection result peaks at $\eta=0.3$ (43.6 AP on COCO-val, 43.0 AP on LVIS-minival). A moderate margin ($\eta=0.1-0.3$) outperforms the no margin setting ($\eta=0$), confirming that our enforced embedding separation enhances discriminative power. However, excessive margins ($\eta\geq0.5$) degrade performance, suggesting overly aggressive separation harms embedding representations. The peak at $\eta=0.3$ balances discriminative features and semantic relationships, validating our choice of $\eta$ in the NNH loss.

\noindent \textbf{Impact of Negative Example Quantity on Model Performance.} As shown in Fig.~\ref{fig:ab_hyperparameters}c, our ablation study shows that employing three negative prompts yields optimal zero-shot detection accuracy (43.6 AP on COCO-val, 43.0 AP on LVIS-minival), representing a 0.6 AP improvement over the single-negative baseline. However, further increasing to five negative examples yields diminishing returns (43.5 AP on COCO-val), suggesting an upper bound on the benefits of negative prompt diversity. This pattern indicates that while multiple negative examples help discriminate between visually similar categories, excessive negative prompts may introduce noise or redundant information.

\noindent \textbf{Impact of Positive Prompt Quantity on Model Performance.} Our ablation study reveals a counterintuitive relationship between the number of positive visual prompts and zero-shot detection performance. As shown in Fig.~\ref{fig:ab_hyperparameters}d, the results indicate that using a single positive example yields optimal performance (43.6 AP on COCO-val, 43.0 AP on LVIS-minival), with progressively degraded results as more examples are incorporated. Specifically, increasing the number of positive examples to 4, 8, and 16 leads to performance drops of 0.6, 0.9, and 0.8 AP points, respectively, on COCO-val, with similar degradation patterns observed on LVIS-minival. This suggests that \ourmodel{} benefits more from high-quality positive examples rather than quantity, potentially due to reduced noise in the learned representations.

\noindent \textbf{Visualization of \ourmodel{}'s three inference modes.}
Fig.~\ref{fig:ab_visulization}a demonstrates \ourmodel{}'s positive-only baseline (using muffin prompts) yields imperfect results, missing some muffins and misclassifying chihuahuas. With Auto-suggested mode (Fig.~\ref{fig:ab_visulization}b), \ourmodel{} automatically synthesizes multiple negative visual prompts conditioned on the positive prompt, generating refined detections with fewer errors. Most impressively, for user-curated mode (Fig.~\ref{fig:ab_visulization}c), \ourmodel{} leverages both user-provided positive and negative prompts to achieve optimal performance—correctly detecting all muffins while eliminating all chihuahua misclassifications. This shows \ourmodel{}'s flexible visual prompt settings and progressive improvement.

\section{Conclusion}
We have introduced \ourmodel{}, a novel framework that advances open-set object detection through the integration of negative visual prompts. Addressing a critical limitation of existing positive-only paradigms—their vulnerability to visually similar distractors, \ourmodel{} achieves enhanced detection performance via three key contributions: (1) a training-free NNC module for hard negative suppression, (2) an NNH loss for embedding space regularization, and (3) a unified architecture for joint positive-negative prompt processing. Extensive experiments demonstrate significantly enhanced robustness, evidenced by both a reduced performance gap between visual and text prompts and superior performance in long-tailed scenarios.

\section{Acknowledgement} This work is partially supported by the National Natural Science Foundation of China (No. 62206068).
\bibliography{aaai2026.bib}

\clearpage

\section{Appendix}
\section{A. Additional Related Work}
\begin{figure*}[t]
\centering
\includegraphics[width=0.8\textwidth]{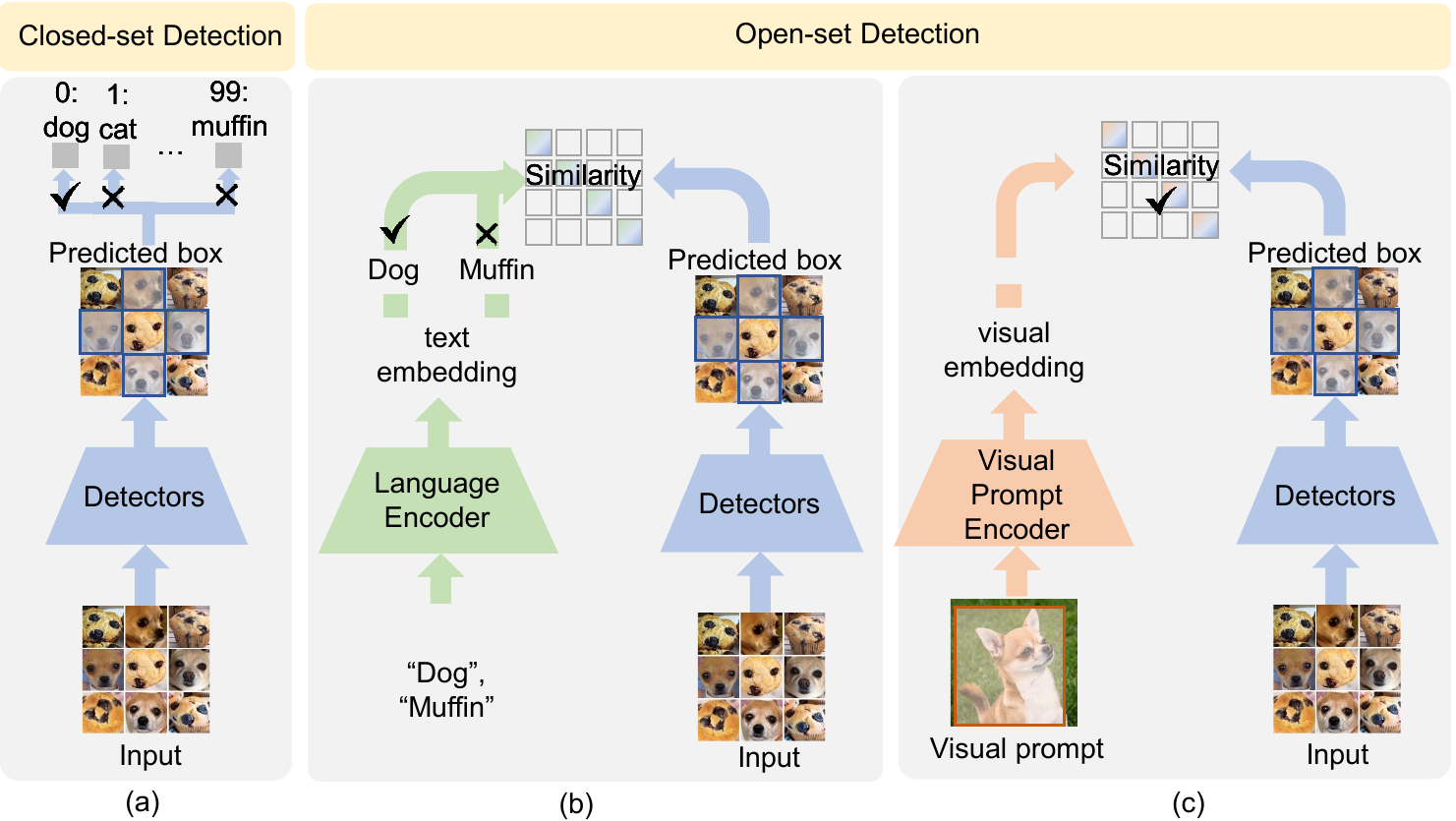}
\caption{Comparison of Object Detection Paradigms.(a) Closed-set object detection methods; (b) Text-prompt object detection methods; (c) Visual-prompt object detection methods.}
\label{fig:Related Work}
\end{figure*}

Figure~\ref{fig:Related Work} illustrates the evolution of object detection paradigms, highlighting three key approaches:

\begin{itemize}
\item \textbf{Closed-Set Detection:} Traditional detectors~\cite{carion2020end,girshick2015fast,li2022dn,lin2017focal,liu2022dabdetr,ren2015faster,zhang2022dino,zhu2020deformable} excel within predefined categories (e.g., COCO-80) but fail on novel objects, as shown in Fig.~\ref{fig:Related Work} (a) where "muffin" predictions occur despite visual dissimilarity to dogs. Retraining is required for new categories.

\item \textbf{Open-set Text-Prompt Detection:} While VLMs like CLIP~\cite{radford2021learning} or BERT~\cite{devlin2019bert} enable open-vocabulary queries in Fig.~\ref{fig:Related Work} (b), Open-set Text-Prompt ~\cite{gu2021open,li2022grounded,liu2024grounding,minderer2022simple,yao2022detclip} rely on text embeddings that struggle with (i) visually similar categories (dog/muffin) that cannot be articulated clearly and (ii) rare objects with inadequate image-text paired data~\cite {gu2021open}.

\item \textbf{Visual-Prompt Open-set Detection:} Visual-prompt approaches~\cite{minderer2022simple,xu2023multi,zang2022open,minderer2022simple,xu2023multi,zang2022open} in Fig.~\ref{fig:Related Work} (c) provide an example visual prompt (box or point) for fine-grained recognition. However, existing implementations focus solely on positive examples, making them susceptible to hard negatives that are visually similar.
\end{itemize}

\section{B. Additional Technical Details}

\noindent \textbf{Visual Prompt Generation.} 
Considering that there currently exists no dedicated dataset for negative visual prompts, and that annotating such datasets is both labor-intensive and time-consuming, we utilize the augmentation of sampled Ground Truth (GT) bounding boxes to construct positive and negative visual prompts. This design is also inspired by the setting of NP-RepMet~\cite{yang2020restoring}, which employs the Intersection over Union (IoU) between GT and Region Proposal Network (RPN) proposals to define positive and negative samples. 

\begin{figure}[h]
\centering
\includegraphics[width=0.45\textwidth]{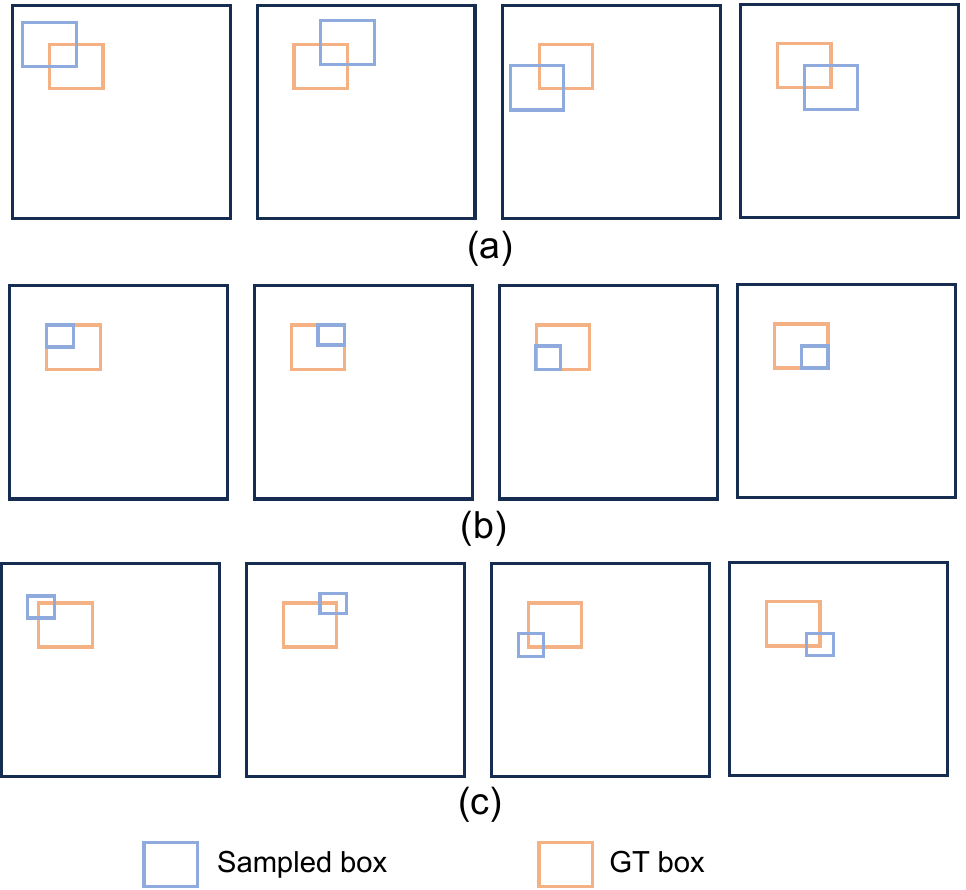}
\caption{Visual Prompt Generation Process. (a) Randomly shift center; (b) Randomly shift size; (c) Randomly shift size and shift center.}
\label{fig:Sampling_mechanism}
\end{figure}

Specifically, for each category $c$ present in an image, we first randomly sample one ground truth bounding box, denoted as $G_{c} = (x_{c}, y_{c}, w_{c}, h_{c})$. Then, positive visual prompts $p_{c}$ are synthesized by applying mild transformations, namely, random scaling or shifting or both within $[0,0.3]$ scale range for the sampled ground truth bounding box $G_{c}$. Similarly, multiple negative visual prompts $n_{c}^{i}, i \in {1,2,...,K}$ are generated with stronger transformations within $[0.7, 1.0]$ scale range by repeating this process $K$ times. Fig.~\ref{fig:Sampling_mechanism} demonstrates the random scaling or shifting process of the sampled ground truth bounding box $G_{c}$.

\noindent \textbf{Implementation Details.}
For Hungarian matching, we employ three key losses: classification loss $\mathcal{L}_{cls}$, box L1 loss $\mathcal{L}_{L1}$, and GIOU loss $\mathcal{L}_{GIoU}$. The loss weights are 2.0, 5.0, and 2.0, respectively. During final training, the weights for $\mathcal{L}_{cls}$, $\mathcal{L}_{Hinge}$, $\mathcal{L}_{L1}$, $\mathcal{L}_{GIoU}$ and $\mathcal{L}_{ DN}$ are set to be 1.0, 1.0, 5.0, 2.0, and 1.0, respectively. All experiments use automatic mixed precision (AMP) for accelerated training and are conducted on 8× NVIDIA A100 GPUs.

\noindent \textbf{Datasets.} The COCO (Common Objects in Context) dataset serves as a fundamental benchmark in object detection research, comprising 80 carefully selected common object categories that represent everyday scenes. In contrast, the LVIS dataset presents a significantly more challenging long-tailed recognition scenario with its extensive vocabulary of 1,203 categories. These categories are explicitly divided into three frequency groups: frequent (405 categories in LVIS-Val, 389 in LVIS-minival), common (461 in LVIS-Val, 345 in LVIS-minival), and rare (337 in LVIS-Val, 70 in LVIS-minival) ~\cite{kamath2021mdetr}. This hierarchical organization enables systematic evaluation of model performance across different levels of category frequency.

Moving beyond these standard benchmarks, the ODinW35 and Roboflow100 datasets provide even broader domain coverage. ODinW35 aggregates 35 distinct datasets, while Roboflow100 expands this diversity with 100 specialized datasets. These comprehensive benchmarks encompass an extensive range of visual domains, including, but not limited to, aerial imagery (e.g., satellite and drone photography), underwater environments, gaming screenshots, and document analysis scenarios. Importantly, all these datasets exhibit pronounced long-tailed distributions, mirroring real-world data imbalance challenges and providing rigorous testbeds for evaluating model robustness across diverse and imbalanced category distributions.


\section{C. Additional Experiment Results}

\begin{table}[h]
\centering
\setlength{\tabcolsep}{1.8mm}
\resizebox{0.47\textwidth}{!}{
\begin{tabular}{c|c|c}
\hline
\textbf{Method} & \textbf{FSC147 test} & \textbf{FSCD-LVIS test} \\
\hline
 & MAE $\downarrow$ & AP $\uparrow$  \\
\hline
FamNet \cite{ranjan2021learning} & 22.08 & - \\
Counting-DETR \cite{nguyen2022few} & - & 22.66 \\
BMNet+ \cite{shi2022represent} & 14.62 & - \\
CountTR \cite{liu2022countr} & 11.95 & - \\
T-Rex \cite{jiang2023t} & \textbf{8.72} & 40.32 \\
T-Rex2 \cite{jiang2024t} & 10.94 & \textbf{43.35} \\
T-Rex-Omni (ours) & 13.76 & 43.27 \\
\hline
\end{tabular}
}
\caption{Few-shot object counting results on FSC147 \cite{ranjan2021learning} and FSCD-LVIS \cite{nguyen2022few} datasets.}
\label{tab:counting_results}
\end{table}

\noindent\textbf{Few-shot Object Counting Results.} We evaluate \ourmodel{} on the few-shot object counting task, where each test image is provided with three visual exemplar boxes of the target object to predict the object count. We evaluate on the FSC147 \cite{ranjan2021learning} and FSCD-LVIS \cite{nguyen2022few} datasets with densely populated small objects. Specifically, FSC147 typically focuses on single-target scenes (one object type per image) with dense, small objects, whereas FSCD-LVIS mainly features multi-target scenes, requiring detection and counting of diverse objects in cluttered environments. We report the Mean Average Error (MAE) metric for FSC147 and the AP metric for FSCD-LVIS, aligning with prior work \cite{jiang2023t,jiang2024t}. The visual exemplar boxes serve as interactive prompts to guide counting. As shown in Tab. \ref{tab:counting_results}, \ourmodel{} achieves competitive performance against state-of-the-art methods. \ourmodel{} (13.76 MAE) outperforms earlier baselines like FamNet (22.08) and BMNet+ (14.62), though it lags behind T-Rex (8.72), suggesting strong few-shot adaptability. \ourmodel{} (43.27 AP) nearly matches T-Rex2 (43.35), demonstrating robustness in multi-target detection and counting. Its performance surpasses Counting-DETR (22.66) by a large margin, highlighting advantages in leveraging visual prompts.


\section{D. Additional Ablation Study}
\noindent\textbf{Impact of Stochastic Mode-switching Probability $B$.} As shown in Tab.\ref{ab:stochastic mode-switching probability}, the ablation study reveals a clear optimal range for the stochastic mode-switching probability $B$, with peak performance achieved at $B(0.5)$ (43.6 AP on COCO-val, 43.0 AP on LVIS-minival). This represents a significant improvement over both extreme settings - +1.8 AP over always-off ($B(0)$) and +1.2 AP over always-on ($B(1)$) configurations on COCO-val. The results demonstrate that moderate stochasticity during training ($B=0.2-0.5$) substantially enhances model generalization, while higher probabilities ($B\geq0.8$) yield diminishing returns. This pattern suggests that balanced exposure to both positive-only and positive-negative training modes is crucial for developing robust feature representations that transfer effectively to the fixed inference setting ($B=1$).

\begin{table}[]
\centering
\setlength{\tabcolsep}{6mm}
\resizebox{0.38\textwidth}{!}{
\begin{tabular}{ccc}
\hline
\begin{tabular}[c]{@{}c@{}}Training\\ Settings\end{tabular} &
  \begin{tabular}[c]{@{}c@{}}COCO-val\\ Zero-shot\end{tabular} &
  \begin{tabular}[c]{@{}c@{}}LIVIS-minival\\ Zero-shot\end{tabular} \\ \hline
$B(0)$   & 41.8 & 40.6 \\
$B(0.2)$ & 42.9 & 41.7 \\
$B(0.5)$ & 43.6 & 43.0 \\
$B(0.8)$ & 43.0 & 41.8 \\
$B(1)$   & 42.4 & 40.9 \\ \hline
\end{tabular}}
\caption{Ablation study on stochastic mode-switching probability $B$ during training. During evluation, $B$ is set to 1 for auto-suggest inference mode.}
\label{ab:stochastic mode-switching probability}
\end{table}

\begin{table}[h]
\centering
\setlength{\tabcolsep}{1.5mm}
\resizebox{0.47\textwidth}{!}{
\begin{tabular}{ccccclc}
\hline
\multicolumn{1}{c|}{Settings} &
  \multicolumn{4}{c|}{Inference Latency (s)} &
  \multicolumn{2}{c}{Frame Rate (1/s)} \\ \cline{2-7} 
\multicolumn{1}{c|}{} &
  Backbone &
  Encoder &
  \begin{tabular}[c]{@{}c@{}}Visual Prompt\\ Encoder\end{tabular} &
  \multicolumn{1}{c|}{Decoder} &
  \multicolumn{1}{c|}{FPS} &
  \begin{tabular}[c]{@{}c@{}}Interactive \\ FPS\end{tabular} \\ \hline
\multicolumn{7}{c}{Swim-T}                                                                                             \\ \hline
\multicolumn{1}{c|}{K=1} & 0.0318 & 0.0240 & 0.0220 & \multicolumn{1}{c|}{0.0180} & \multicolumn{1}{l|}{10.45} & 25.00 \\
\multicolumn{1}{c|}{K=3} & 0.0318 & 0.0240 & 0.0435 & \multicolumn{1}{c|}{0.0180} & \multicolumn{1}{l|}{8.55}  & 16.26 \\
\multicolumn{1}{c|}{K=5} & 0.0318 & 0.0240 & 0.0644 & \multicolumn{1}{c|}{0.0180} & \multicolumn{1}{l|}{7.24}  & 12.13 \\ \hline
\multicolumn{7}{c}{Swim-L}                                                                                             \\ \hline
\multicolumn{1}{c|}{K=1} & 0.0318 & 0.0240 & 0.0475 & \multicolumn{1}{c|}{0.0180} & \multicolumn{1}{l|}{8.24}  & 15.27 \\
\multicolumn{1}{c|}{K=3} & 0.0318 & 0.0240 & 0.0943 & \multicolumn{1}{c|}{0.0180} & \multicolumn{1}{l|}{5.95}  & 8.91  \\
\multicolumn{1}{c|}{K=5} & 0.0318 & 0.0240 & 0.1357 & \multicolumn{1}{c|}{0.0180} & \multicolumn{1}{l|}{4.78}  & 6.51  \\ \hline
\end{tabular}
}
\caption{Inference latency and frame rates for different numbers of negative prompts $K$. Interactive FPS measures the prompt encoder and decoder, which are re-run for each interaction in the late-fusion architecture, while the backbone and encoder are executed once.}
\label{ab:inference cost}
\end{table}

\noindent\textbf{Model Efficiency Analysis under Different $K$ Negative Visual Prompts.} This section evaluates the computational efficiency of \ourmodel{} with respect to the number of negative visual prompts ($K$). Our experiments, conducted on an NVIDIA RTX 3090 GPU under the same settings as T-Rex2~\cite{jiang2024t}, demonstrate that increasing $K$ only linearly impacts the inference time of the visual prompt branch—growing from 0.0220s ($K=1$) to 0.0644s ($K=5$) for Swim-T—while the latency of other components remains unchanged. Critically, this linear scaling results in only a moderate reduction in frame rate, allowing \ourmodel{} to maintain interactive speeds of 6.51–12.13 FPS across all tested configurations. These results confirm that our approach remains suitable for real-time applications even as the number of negative prompts increases.

\section{E. Additional Visualization Results}
\begin{figure*}[t]
\centering
\includegraphics[width=0.9\textwidth]{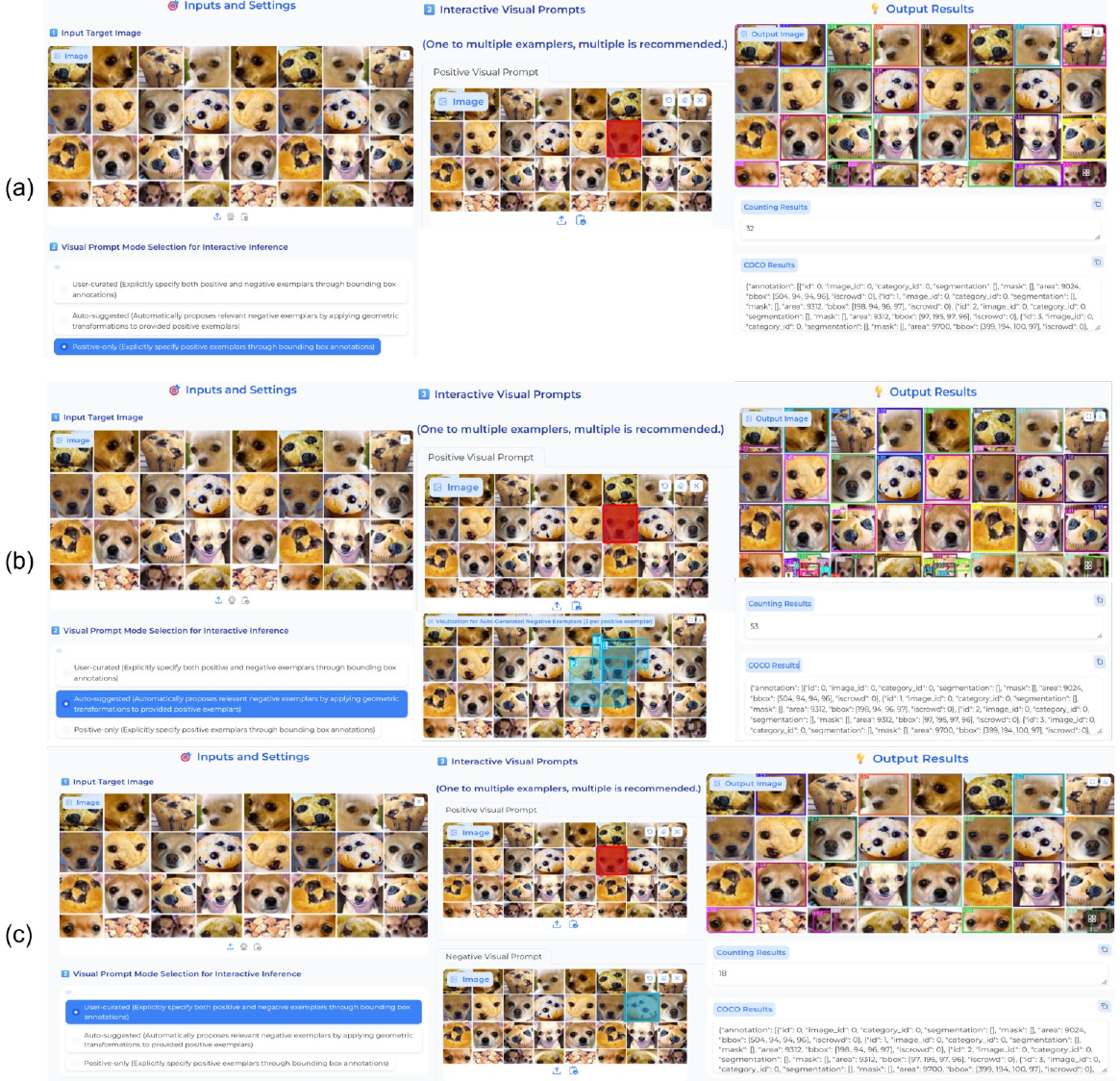}
\caption{Additional visualization detection results for visual prompts within the input image. We present \ourmodel{} under three inference modes: (a) Positive-only; (b) Auto-suggested; (c) User-curated.}
\label{fig:Demo1}
\end{figure*}

\begin{figure*}[t]
\centering
\includegraphics[width=0.6\textwidth]{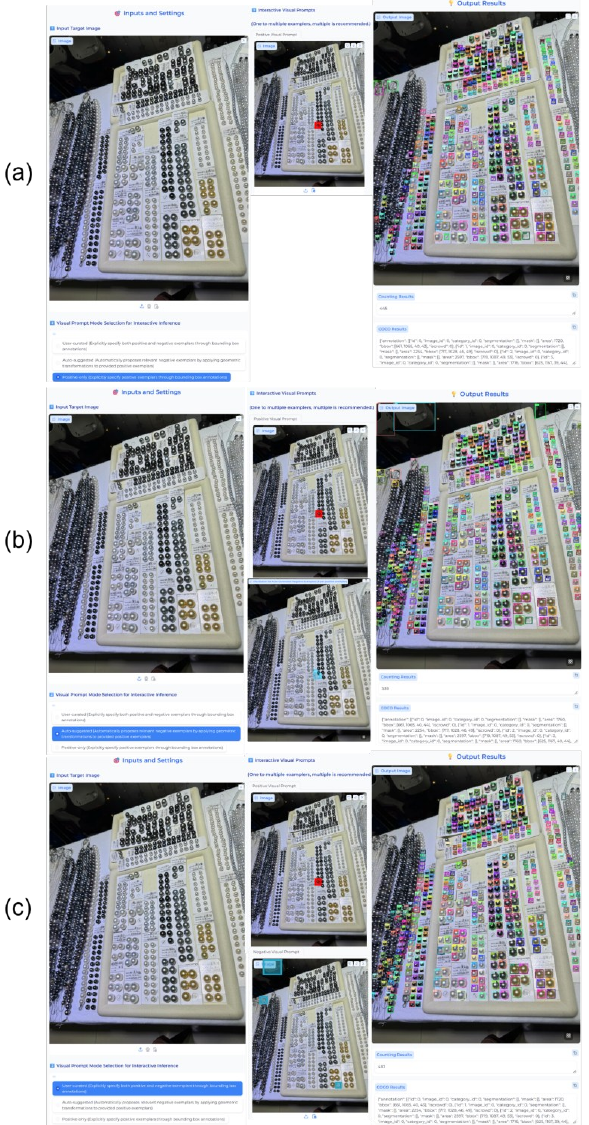}
\caption{Additional visualization detection results for visual prompts within the input image. We present \ourmodel{} under three inference modes: (a) Positive-only; (b) Auto-suggested; (c) User-curated.}
\label{fig:Demo2}
\end{figure*}

\begin{figure*}[t]
\centering
\includegraphics[width=0.9\textwidth]{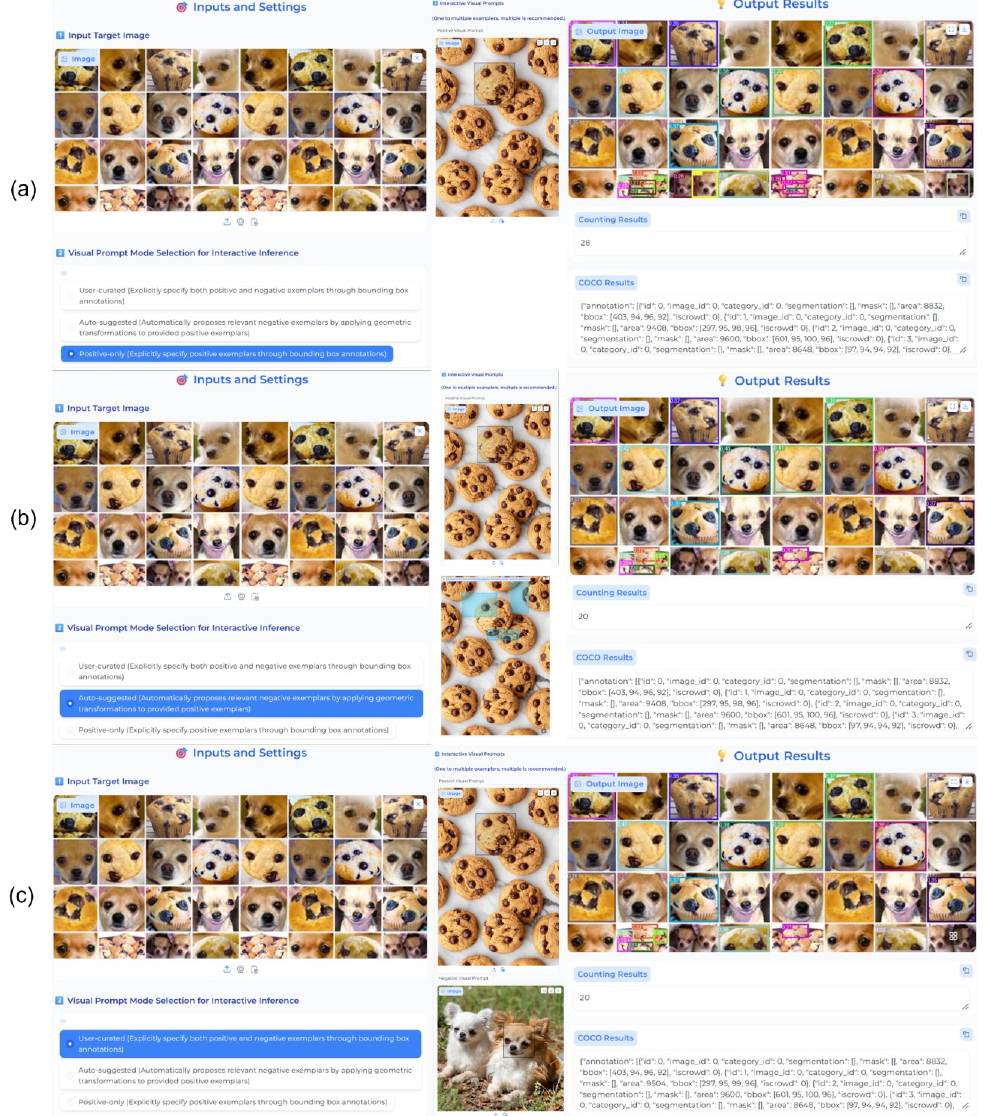}
\caption{Additional visualization detection results for visual prompts cross the other image. We present \ourmodel{} under three inference modes: (a) Positive-only; (b) Auto-suggested; (c) User-curated.}
\label{fig:Demo3}
\end{figure*}

\begin{figure*}[t]
\centering
\includegraphics[width=0.7\textwidth]{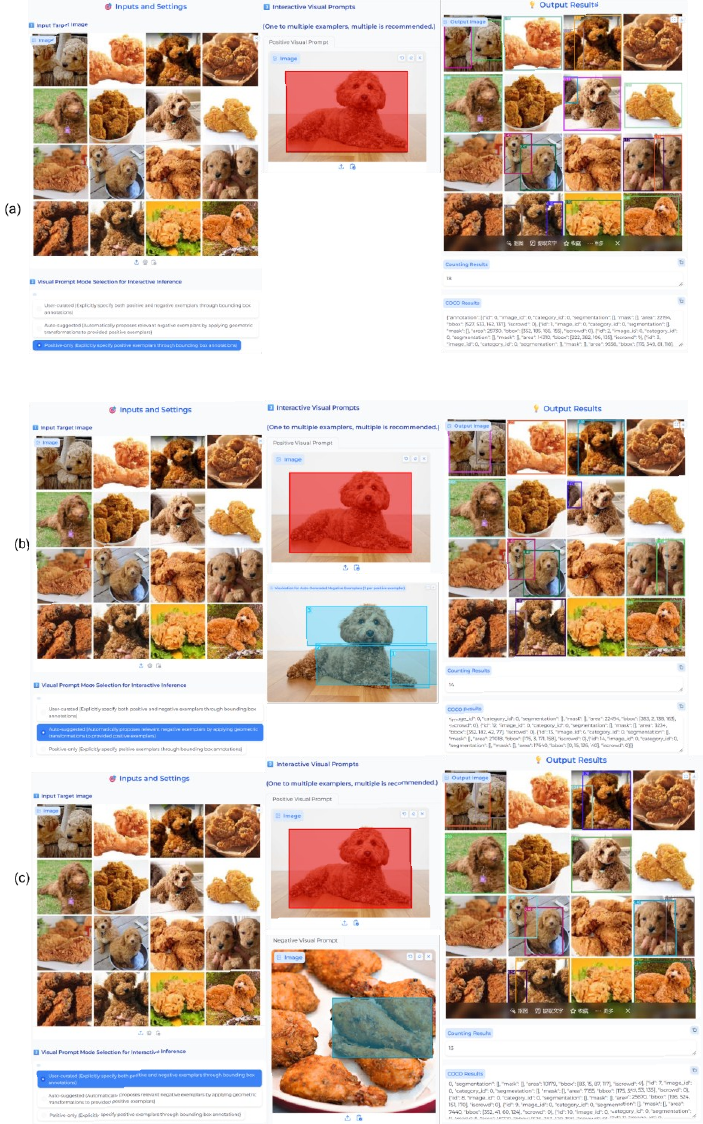}
\caption{Additional visualization detection results for visual prompts cross the other image. We present \ourmodel{} under three inference modes: (a) Positive-only; (b) Auto-suggested; (c) User-curated.}
\label{fig:Demo4}
\end{figure*}

\begin{figure*}[t]
\centering
\includegraphics[width=0.7\textwidth]{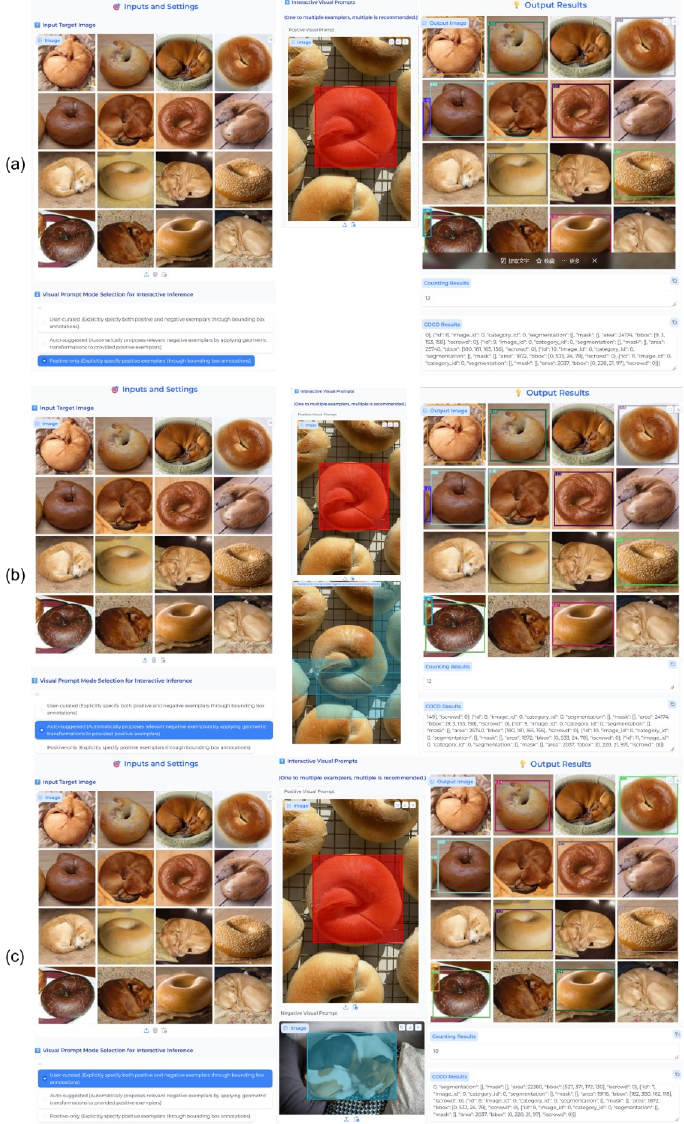}
\caption{Additional visualization detection results for visual prompts cross the other image. We present \ourmodel{} under three inference modes: (a) Positive-only; (b) Auto-suggested; (c) User-curated.}
\label{fig:Demo5}
\end{figure*}

Figs. \ref{fig:Demo1}–\ref{fig:Demo5} present the detection results of the proposed model under corner cases across three inference modes: (a) positive-only, (b) auto-suggested, and (c) user-curated. Specifically, Figs. \ref{fig:Demo1} and \ref{fig:Demo2} illustrate the detection of in-image visual prompts, where the visual prompt is derived from the cropped block of the detected image. In contrast, Figs. \ref{fig:Demo3}–\ref{fig:Demo5} demonstrate the detection of cross-image visual prompts, with the visual prompt sourced from distinct images.

\section{F. Future work.}
This work opens several promising directions: (i) integrating LLM-based negative prompts in visual reasoning for object detection and (ii) applications in safety-critical domains like medical imaging. By bridging the gap between human and machine visual discrimination, \ourmodel{} establishes a new paradigm for robust open-set recognition.
\noindent \textbf{Limitations.} \ourmodel{} exclusively focuses on visual prompts, leaving potential text-visual synergies unexplored. Although being effective for false positive suppression, \ourmodel{} optimizes existing decision boundaries rather than improving rare-object embedding space. 

\clearpage

\end{document}